\documentclass[letterpaper, 10 pt, conference]{ieeeconf}

\IEEEoverridecommandlockouts
\usepackage{cite}
\usepackage{amsmath,amssymb,amsfonts}
\usepackage{algorithmic}
\usepackage{graphicx}
\usepackage{textcomp}
\usepackage{xcolor}
\usepackage{booktabs}
\usepackage{multirow}
\usepackage{verbatim}
\usepackage{hyperref}
\usepackage{lipsum} % For dummy text to show layout
\usepackage{tabularx} % For better table width management
\usepackage{ragged2e} % For better justification in table cells
\usepackage[export]{adjustbox}
\usepackage{booktabs}

\usepackage{color}

\def\BibTeX{{\rm B\kern-.05em{\sc i\kern-.025em b}\kern-.08em
    T\kern-.1667em\lower.7ex\hbox{E}\kern-.125emX}}

\newcolumntype{L}{>{\RaggedRight\arraybackslash}X}

\begin{document}

\title{\LARGE \bf XRoboToolkit: A Cross-Platform Framework for Robot Teleoperation}

\author{Zhigen Zhao$^{1,2}$, Liuchuan Yu$^{1,3}$, Ke Jing$^{1}$ and Ning Yang$^{1}$% <-this % stops a space
\thanks{$^{1}$ByteDance, PICO, San Jose, CA, 95110, USA. Send correspondence to {\tt\small ke.jing@bytedance.com, yangning726@gmail.com}}%
\thanks{$^{2}$Georgia Institute of Technology, Institute for Robotics and Intelligent Machines (IRIM), 
        Atlanta, GA 30332, USA
        {\tt\small zhigen.zhao@gatech.edu}}%
\thanks{$^{3}$George Mason University, Computer Science, Fairfax, Virginia 22030, USA
    {\tt\small lyu20@gmu.edu}}%
}

\maketitle
\thispagestyle{empty}
\pagestyle{empty} 
\begin{abstract}
The rapid advancement of Vision-Language-Action models has created an urgent need for large-scale, high-quality robot demonstration datasets. Although teleoperation is the predominant method for data collection, current approaches suffer from limited scalability, complex setup procedures, and suboptimal data quality. This paper presents XRoboToolkit\footnote[4]{Website: \url{https://xr-robotics.github.io}}, a cross-platform framework for extended reality-based robot teleoperation built on the OpenXR standard. The system features low-latency stereoscopic visual feedback, optimization-based inverse kinematics, and support for diverse tracking modalities, including head, controller, hand, and auxiliary motion trackers. XRoboToolkit's modular architecture enables seamless integration across robotic platforms and simulation environments, spanning precision manipulators, mobile robots, and dexterous hands. We demonstrate the framework's effectiveness through precision manipulation tasks and validate data quality by training VLA models that exhibit robust autonomous performance.
\end{abstract}

\section{Introduction}
Recent advances in deep generative robot learning~\cite{urain2024deep}, especially Vision-Language-Action Models (VLAs)~\cite{sapkota2025vision, black2024pi_0, kim2024openvla, team2025gemini}, are critically dependent on large-scale, high-quality datasets containing robot skill demonstrations. Robot teleoperation~\cite{si2021review, darvish2023teleoperation, li2025train, buckley2024dexterous, iyer2025open} is one of the primary approaches for generating human demonstrations of complex manipulation and mobility tasks, leveraging human operators' natural ability to generalize across diverse environments and tasks.

Recent robot teleoperation frameworks follow several paradigms, each with distinct trade-offs. Leader-follower approaches~\cite{zhao2023learning} offer low latency and intuitive operation but require custom hardware tailored to specific robot platforms, limiting scalability and accessibility. Vision-based teleoperation systems~\cite{qin2023anyteleop} provide greater flexibility and generalizability across diverse robotic hardware but often suffer from unstable tracking performance and higher latency, degrading operator performance and data quality. Virtual Reality (VR) or Extended Reality (XR) teleoperation~\cite{wang2024towards, chengopen, OrbikEbert2021OculusReader, seo2023deep} has emerged as a promising alternative, utilizing commercially available headsets to create intuitive control interfaces with stereoscopic visual feedback that generalize across multiple platforms. However, existing XR solutions remain difficult to configure and often rely on individual Unity SDKs or WebXR platforms that introduce additional latency and compatibility challenges. Another significant limitation is the lack of standardized data formats between XR devices and robot controllers, necessitating substantial integration work for new XR devices or robot platforms.

To address these limitations, we present XRoboToolkit—a comprehensive suite of cross-device software development kits and applications for real-time robot teleoperation via XR devices. The toolkit provides a generalized interface layer that resolves standardization challenges by adopting OpenXR~\cite{openxr_spec} conventions on the XR side and modular, extensible Python and C++ interfaces on the robot side for seamless integration across robotic platforms. Current support includes devices such as PICO 4 Ultra and Meta Quest 3.

A major contribution is the stereoscopic visual feedback system, which integrates a low-latency communication protocol and a highly efficient video streaming pipeline, both optimized to minimize latency and reduce motion sickness. 

On the robot side, the system employs a quadratic programming (QP)-based inverse kinematics (IK) solver that generates smooth, reliable robot motion, particularly near kinematic singularities, and incorporates dexterous hand tracking for retargeting human hand motions to robotic hands in fine-grained manipulation tasks. The modular architecture of XRoboToolkit enables straightforward integration with diverse robotic systems and simulation environments, tested on platforms such as UR5 and ARX R5 arms, the Galaxea R1-Lite mobile manipulator, the Shadow dexterous hands, with native support for MuJoCo~\cite{todorov2012mujoco}.

Sec.~\ref{sec:system} details the architecture of the XRoboToolkit. Example applications are presented in Sec.~\ref{sec:example}, performance evaluation in Sec.~\ref{sec:experiments}, and conclusions in Sec.~\ref{sec:conclusions}.

\section{Teleoperation System} \label{sec:system}

\subsection{Overview}

\begin{figure*}[!ht]
    \centering
    \includegraphics[width=\textwidth]{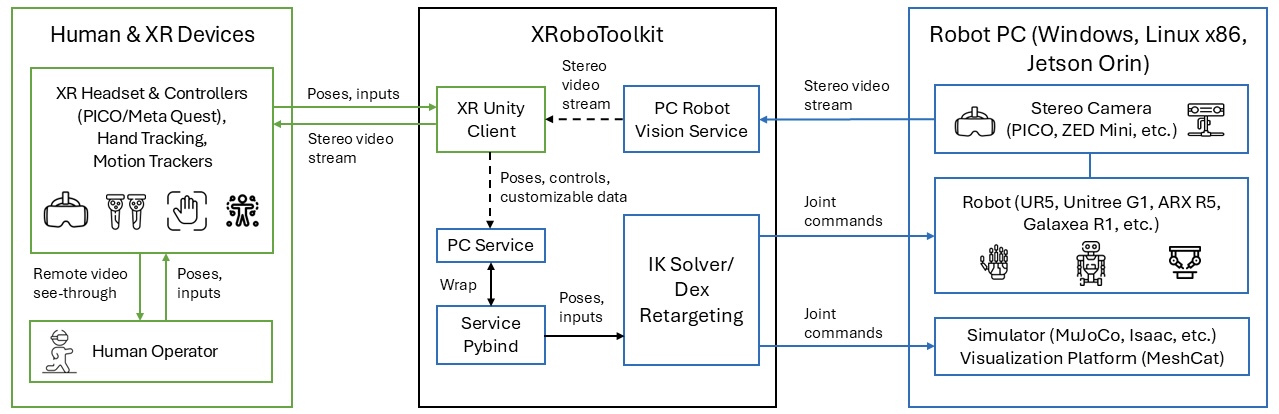}
    \caption{Overview of \texttt{XRoboToolkit}, an integrative framework bridging XR and robotics. Core functionalities include real-time teleoperation and stereoscopic vision. Green blocks represent XR-side components, while blue blocks indicate components on the robot side, which runs on either the robot PC or a separate PC connected to the same network as the headset.
    }
    \label{fig:overview}
    \vspace{-0.2in}
\end{figure*}

Fig.~\ref{fig:overview} presents an overview of the XRoboToolkit architecture. 
The \texttt{Unity Client} application, deployed on XR headsets, captures pose tracking data and delivers a stereoscopic visual interface for the human operator. This pose tracking data, including head, hand, controller, full-body, and object tracking (via motion trackers), is transmitted to the \texttt{PC Service} in C++, the specific tracking data format is discussed in Sec.~\ref{sec:data_streaming}. Additionally, the package \texttt{PC Service Pybind} allows direct access to the XR tracking data in Python without handling the raw data structure. Stereo vision is enabled either through the onboard cameras of PICO headsets or an external ZED Mini camera with the module \texttt{Robot Vision}. The \texttt{IK} and \texttt{Dexterous Retargeting} solvers are implemented in the robot teleoperation module to provide support for both simulated environments, such as MuJoCo, and physical robot platforms, including the UR5, ARX R5, and Galaxea R1-Lite. The XRoboToolkit's modular architecture facilitates easy integration with additional simulators and robotic platforms, providing a flexible and extensible solution for stereoscopic teleoperation in virtual and physical environments.

\subsection{Data Streaming}
\label{sec:data_streaming}
\texttt{XRoboToolkit-PC-Service} employs an asynchronous, callback-driven architecture for real-time data streaming from VR hardware to client applications. Communication is managed via a dedicated SDK handling connection to the streaming service and data payload reception.

% \texttt{PC Service} employs an asynchronous, callback-driven architecture for real-time data streaming from VR hardware to client applications. Communication is managed via a dedicated SDK handling connection to the streaming service and data payload reception.

\begin{figure}
    \centering
    \includegraphics[width=0.9\linewidth]{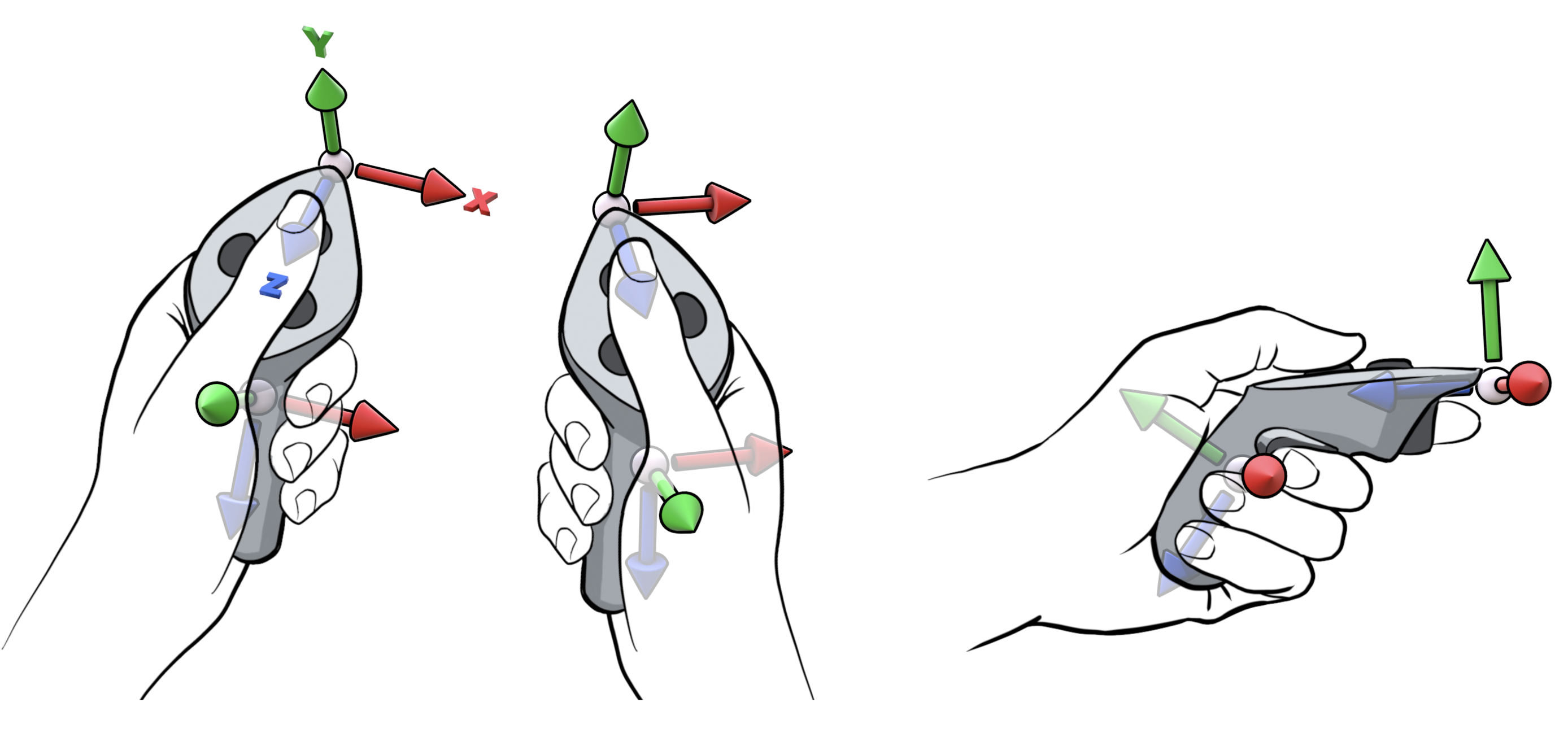}
    \caption{OpenXR conventions for pose tracking coordinate system~\cite{openxr_spec}.}
    \label{fig:openxr_conventions}
    \vspace{-0.2in}
\end{figure}

\textbf{XR Data Formats: }
Following OpenXR conventions, all positional and rotational data use a right-handed coordinate system with X-axis right, Y-axis up, and Z-axis backward, as shown in Fig.~\ref{fig:openxr_conventions}(a). The origin is established at the user's head position when the application launches. The 6 degree-of-freedom (DOF) pose data are formatted as seven floating-point numbers separated by commas: 3D position vector $[x, y, z]$ followed by quaternion $[qx, qy, qz, qw]$.

All real-time tracking data are transmitted within a single JSON object at 90 Hz. This design simplifies client-side parsing and ensures a consistent data structure regardless of enabled tracking features. Table~\ref{tab:xr_data_formats} provides an overview of the main tracking data fields in XRoboToolkit.

\begin{table}[!ht]
\centering
\caption{XR Tracking Data Formats}
\label{tab:xr_data_formats}
\footnotesize
\begin{tabularx}{\columnwidth}{@{}llL@{}}
\toprule
\textbf{Type} & \textbf{Field} & \textbf{Description} \\ 
\midrule
\multirow{4}{*}{\textbf{Head}} 
& \texttt{pose} & Headset pose \\
& \texttt{status} & Tracking confidence (0: unreliable, 1: reliable) \\
& \texttt{handMode} & Input mode (0: None, 1: controller, 2: hand) \\
\midrule
\multirow{9}{*}{\textbf{Controller}} 
& \texttt{pose} & Controller pose \\
& \texttt{axisX, axisY} & Joystick position \\
& \texttt{axisClick} & Joystick press state \\
& \texttt{grip} & Grip input \\
& \texttt{trigger} & Trigger input \\
& \texttt{primaryButton} & X Button, A Button \\
& \texttt{secondaryButton} & Y Button, B Button \\
& \texttt{menuButton} & Menu (L), Screenshot/Record (R) \\
\midrule
\multirow{3}{*}{\textbf{Hand}} 
& \texttt{isActive} & Tracking status (0: not active, 1: active)\\
& \texttt{scale} & Hand scale factor \\
& \texttt{HandJointLocations} & Array of 26 hand joint data entries \\
\midrule
\multirow{1}{*}{\textbf{Whole-Body}} 
& \texttt{joints} & Array of 24 body joint data entries\\
\midrule
\multirow{4}{*}{\textbf{Motion Tracker}} 
& \texttt{p} & Pose \\
& \texttt{va} & Velocity \& angular velocity \\
& \texttt{wva} & Acceleration \& angular acceleration \\
& \texttt{sn} & Unique serial number of tracker \\
\bottomrule
\end{tabularx}
\vspace{-0.2in}
\end{table}

\textbf{Head Tracking: }
Head tracking data contains headset pose, status integer indicating tracking confidence, and hand mode integer specifying active input mode.

\textbf{Controller Tracking: }
Controller tracking captures both left and right controllers' poses with button and joystick states. Joystick axes \texttt{axisX} and \texttt{axisY} provide floating-point values from -1 to 1. \texttt{grip} and \texttt{trigger} inputs are analog controls with values between 0 and 1 indicating pressure intensity. The remaining buttons provide binary state information.

\begin{figure}
    \centering
    \includegraphics[width=\linewidth]{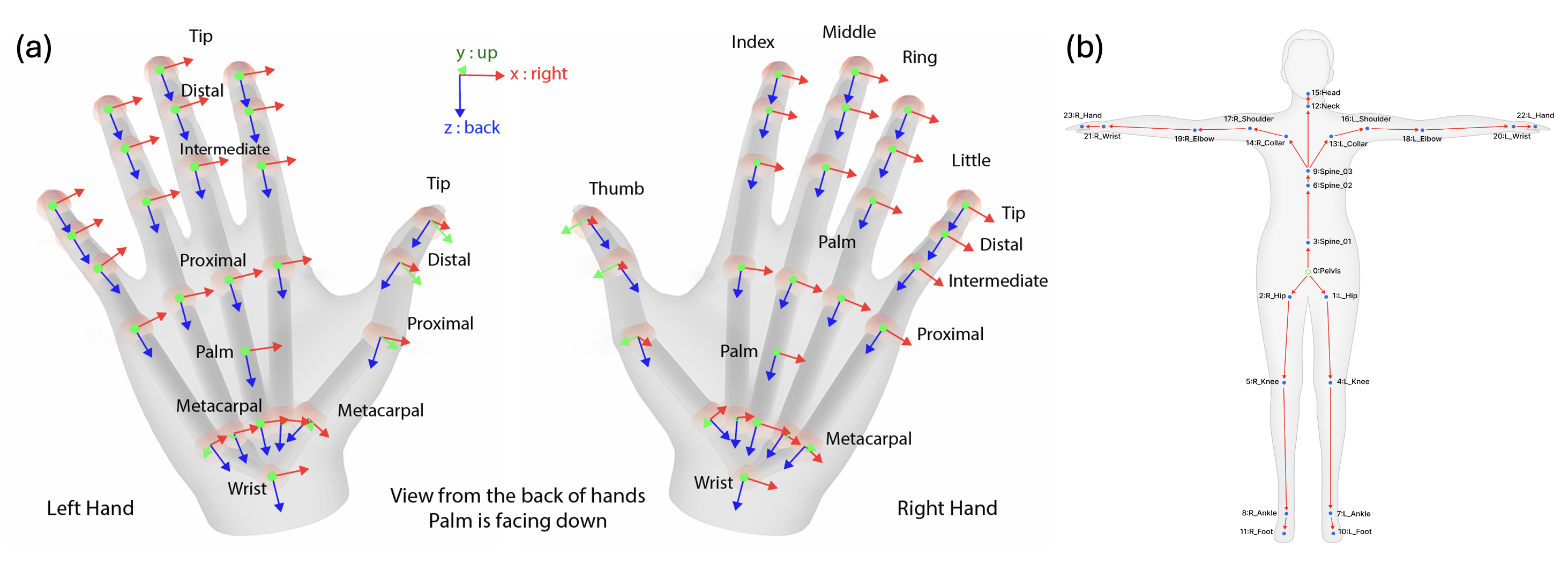}
    \caption{Conventions for (a) hand tracking keypoints and (b) whole-body tracking keypoints.}
    \label{fig:hand_and_body}
    \vspace{-0.2in}
\end{figure}

\textbf{Hand Gesture Tracking: }
Each hand gesture is represented through 26 joint poses~\cite{openxr_spec}: 4 joints on the thumb, 5 joints on each remaining finger, plus palm and wrist joints, as illustrated in Fig.~\ref{fig:hand_and_body}(a). Hand tracking data includes tracking quality, scale factor, and an array of 26 keypoint data entries per hand. Each entry contains a 6-DOF pose with additional metadata, including tracking status and joint radius. While transmitted with the JSON object at 90 Hz, hand tracking data updates at 60 Hz due to camera limitations.

\textbf{Whole-Body Motion Capture: }
Whole-body joint tracking consists of 24 keypoint data entries corresponding to major human model joints, as shown in Fig.~\ref{fig:hand_and_body}(b). Each entry contains joint pose, velocity, and acceleration. The 24-joint model follows PICO's standard, as OpenXR currently lacks a standardized whole-body model.

\textbf{Motion Tracker:} 
For PICO 4 Ultra headsets, we support object tracking mode for auxiliary motion trackers. Motion tracker data captures pose, velocity, and acceleration measurements with serial numbers for tracker identification.

\subsection{Robot Control}
The robot control module maps XR tracking state to robot commands through distinct control modes: IK for manipulator control, dexterous hand retargeting, head tracking, and mobile base control.

\subsubsection{Inverse Kinematics}
\label{sec:IK}
For manipulator control, we implement a QP-based IK solver using PlaCo~\cite{placo}, built on the Pinocchio rigid body dynamics library~\cite{carpentier2019pinocchio}. The QP problem is defined as:
\begin{align}
\label{eq:constraints}
\min_{\dot{\mathbf{q}}}&\sum_{i=1}^{N}\mathbf{w}_i\|\mathbf{J}_i(\mathbf{q})\dot{\mathbf{q}}+\mathbf{e}_i(\mathbf{q}) \|^2\\
    \text{s.t. }& l \leq \mathbf{C}(\mathbf{q})\dot{\mathbf{q}}\leq u, \nonumber
\end{align}
where $\mathbf{q}$ represents manipulator configuration; each task $i$ is defined as residual function $\mathbf{e}_{i}(\mathbf{q})$ with weight $\mathbf{w}_i$; $\mathbf{J}_i(\mathbf{q})$ denotes task Jacobian; and $\mathbf{C}(\mathbf{q})$ represents additional constraint matrix.

The optimization-based IK approach enables easy inclusion of constraints and regularization terms. To improve robot stability near singularities, a regularization term maximizes manipulability~\cite{haviland2020purely}:
\begin{equation}\label{eq:manipulability}
    m = \sqrt{\text{det}(\mathbf{J}(\mathbf{q})\mathbf{J}(\mathbf{q})^\top)}
\end{equation}

When using VR controllers, end-effector tracking activates when the user holds the \texttt{grip} button. For a stable, intuitive experience, the system uses relative motion: the robot's end-effector tracks controller displacement relative to its state when \texttt{grip} was first pressed.

Auxiliary motion trackers can be attached to the user's body (e.g., elbow) to introduce additional pose constraints into the QP-based IK. This enables nuanced control over the robot's whole-body posture by mapping operator positions to equivalent robot links, particularly useful for resolving null-space redundancies and achieving anthropomorphic motions.

\subsubsection{Dexterous Hand Retargeting}
\label{sec:dex_retarget}
For dexterous manipulation tasks, keypoint positions from the OpenXR hand tracking model are obtained via the XR headset's native hand tracking algorithm, as specified in Sec.~\ref{sec:data_streaming}. These keypoints are mapped to robot hand joint space via:
\begin{align}
    \min_{\mathbf{q}_{t}}&\sum_{i=1}^{N}\|\alpha \mathbf{v}_t^i-f_i(\mathbf{q}_{t})\|^2+\beta\|\mathbf{q}_{t}-\mathbf{q}_{t-1}\|^2,\\
    \text{s.t. }&\mathbf{q}_{l}\leq\mathbf{q}_{t}\leq\mathbf{q}_{u}\nonumber,
\end{align}
where $\mathbf{q}_{t}$ is robot hand joint configuration at time $t$, $\mathbf{v}_t^i$ is the $i$-th keypoint position in the human hand model, $f_i(\mathbf{q}_{t})$ computes corresponding robot hand position, $\alpha$ is a scaling factor for different hand sizes, and $\beta$ is regularization weight for smooth motion. Implementation uses \texttt{dex\_retargeting}~\cite{qin2023anyteleop}.

\subsubsection{Mobile Base Control}
For mobile manipulators with omnidirectional platforms, the mobile base is controlled by XR controller joysticks. The left joystick's X and Y axes issue linear velocity commands in the robot's sagittal and coronal planes, while the right joystick's X-axis controls angular velocity, providing an intuitive mobility interface during manipulation tasks.

\subsection{XR Unity Application}
\begin{figure}[!t]
    \centering
    \includegraphics[width=0.95\linewidth]{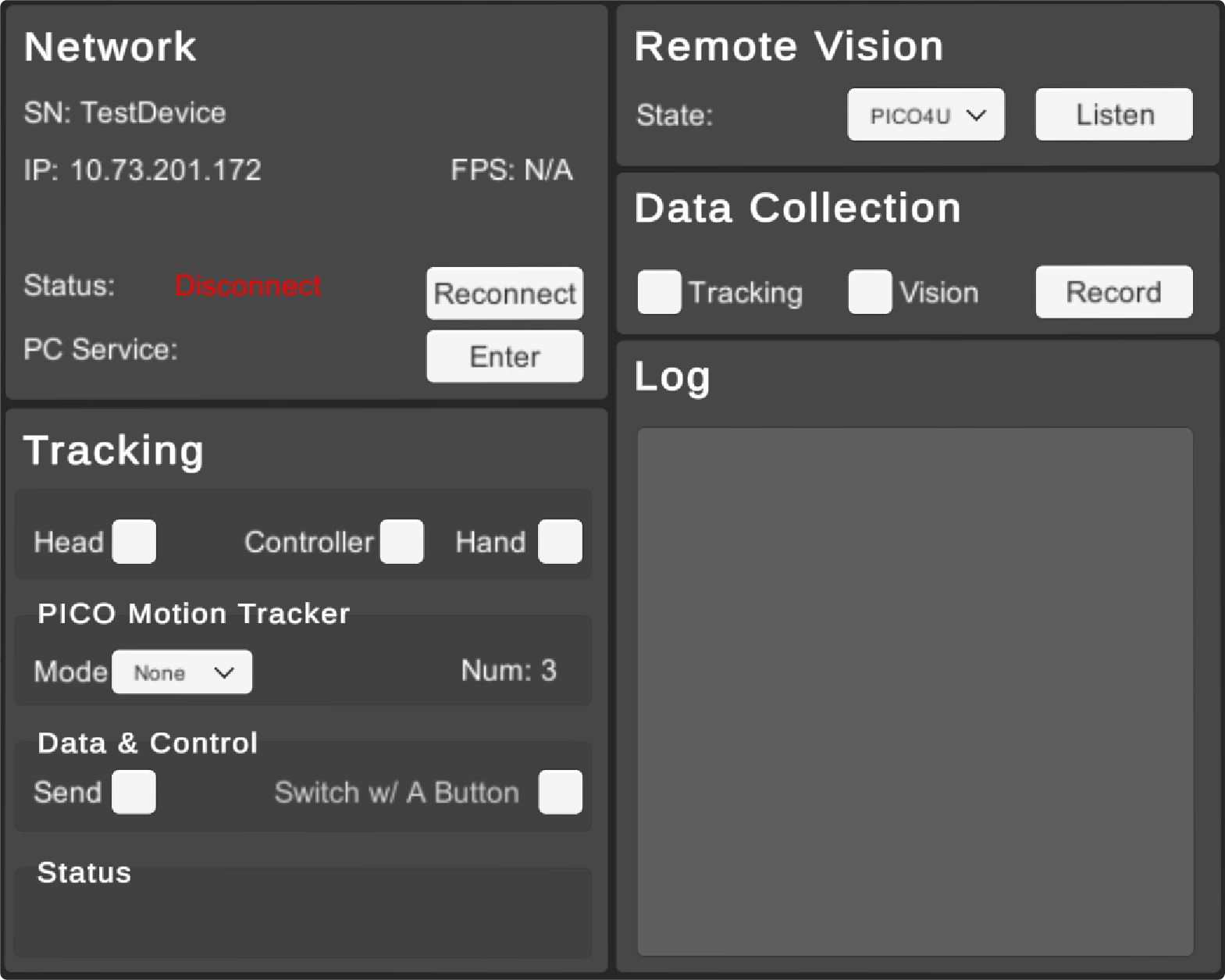}
    \caption{Screenshot of the PICO version of the XR Unity Application.}
    \label{fig:unity_client}
    \vspace{-0.25in}
\end{figure}

Fig.~\ref{fig:unity_client} presents the application interface for \texttt{XRoboToolkit-Unity-Client}, which contains five panels: Network, Tracking, Remote Vision, Data Collection, and Log.

The \textbf{Network panel} displays essential headset status information, including serial number (SN), IP address, frame rate, connection status with the PC service, and configured service IP. The \textbf{Tracking panel} organizes controls into four functional groups. The \texttt{Source} group allows users to select pose data types for tracking (e.g., head, controller, hand). The \texttt{PICO Motion Tracker} group configures tracking modes: None, Full-Body, or Object. Full-Body mode transmits whole-body keypoint poses, while Object mode directly returns motion tracker poses.
Note that for Meta Quest, the auxiliary motion tracker is currently not available. 
The \texttt{Data \& Control} group provides toggles for enabling and disabling all data transmission to the PC service. The \texttt{Status} group provides live tracking status. The \textbf{Remote Vision panel} handles the stereoscopic vision state. The \texttt{State} field displays the current video stream status. The XR application currently supports PICO 4 Ultra and ZED Mini cameras as streaming sources, with extensibility for additional camera sources through modifications of a YAML configuration file. The \textbf{Data Collection panel} enables timestamped recording of both pose and visual data streams on the headset. Note that robot state data is recorded separately through the teleoperation module. The \textbf{Log panel} provides real-time system diagnostics and monitoring information for debugging purposes.

\subsection{Stereoscopic Visual Feedback}

Our toolkit currently supports two stereo video sources: the PICO 4 Ultra headset and the ZED Mini camera. The PICO 4 Ultra operates as a standalone solution, whereas the ZED Mini requires connection to an external computing platform, such as a Windows or Linux PC, or an NVIDIA Orin device, to enable video streaming. When using the PICO 4 Ultra, operators benefit from a homogeneous visual experience, as both display and capture are provided through the same hardware platform.

To achieve stereoscopic vision, we implemented a custom shader that adjusts the interpupillary distance and sets the focal point at approximately 3.3 feet. This configuration provides enhanced three-dimensional depth perception at a distance suitable for manipulation tasks based on our experimental experience, although it comes at the cost of sacrificing far-distance depth accuracy.

Our empirical observations indicate that the PICO 4 Ultra delivers superior visual quality compared to the ZED Mini, particularly in terms of tone reproduction, brightness, color accuracy, and dynamic range. Additionally, the PICO 4 Ultra offers a more balanced field of view (FOV), with horizontal and vertical angles of approximately 76.35° and 61.05°, respectively, compared with the 82° horizontal and 52° vertical FOV of the ZED Mini.

% \begin{itemize}
%     \item PICO 4 Ultra as the stereo camera: 
%     \item ZED Mini
% \end{itemize}
\vspace{-0.05in}
\section{Applications and Demonstrations} \label{sec:example}
\begin{figure*}[!ht]
    \centering
    \includegraphics[width=\textwidth]{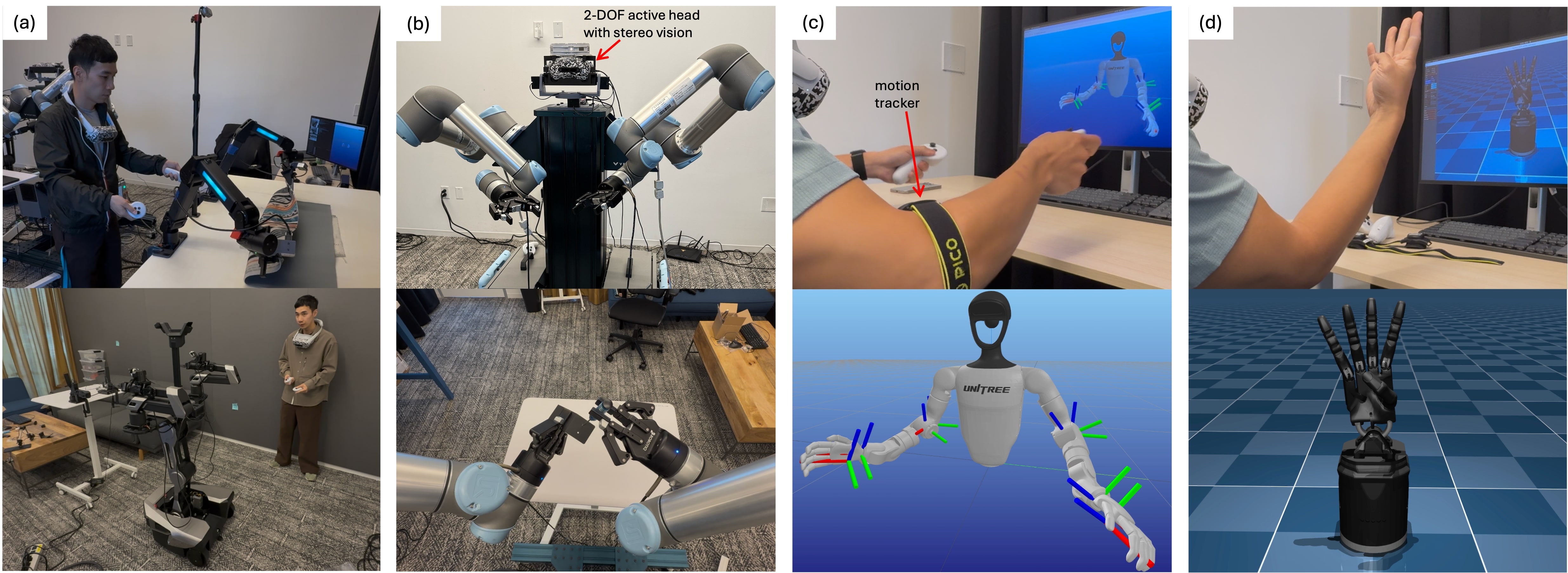}
    \caption{Example applications of XRoboToolkit: (a) teleoperation with XR controllers for dual arm manipulation and mobile manipulators, (b) Dual UR5 manipulators with 2-DOF head tracking and stereo vision, (c) auxiliary motion trackers for robot elbow control in MeshCat visualization, and (d) dexterous hand tracking in MuJoCo simulation.}
    \label{fig:applications}
    \vspace{-0.2in}
\end{figure*}
Fig.~\ref{fig:applications} demonstrates the versatility of XRoboToolkit across diverse robotic platforms and simulation environments. The framework's modular architecture enables seamless integration with both hardware systems and virtual environments, supporting a wide range of teleoperation scenarios from precise manipulation to mobile robotics and dexterous hand control. Please refer to the supplementary video\footnote[5]{Video link: \url{https://youtu.be/g6QJX2s-RCo}} for more information.

% tinyurl: https://tinyurl.com/2m5kme6u

\subsection{XR Controller-Based Teleoperation}
\label{sec:controller_task}
XRoboToolkit supports intuitive teleoperation using XR controllers for both dual-arm manipulation systems and mobile manipulators. The operator wears an XR headset and uses handheld controllers to directly control robot end-effectors through the inverse kinematics solver described in Sec.~\ref{sec:IK}. 
Note that the XR headset can be worn around the neck when stereoscopic visual feedback is not required. This configuration reduces head-mounted weight and fatigue while maintaining full controller tracking functionality, making it particularly suitable for tasks where operators can rely on direct visual observation of the robot workspace.

As shown in Fig.~\ref{fig:applications}(a), the controller-based teleoperation has been validated on multiple platforms, including dual ARX R5 manipulators for long-horizon tasks such as bimanual carpet folding, and the Galaxea R1-Lite mobile manipulator for transportation and placement tasks. The system has also been used in the applications discussed in Sec.~\ref{sec:ur5_task}-\ref{sec:motion_tracker_task}.

\subsection{Precision Manipulation with Active Stereo Vision}
\label{sec:ur5_task}
Fig.~\ref{fig:applications}(b) illustrates our dual UR5 setup equipped with a 2-DOF active head and stereo vision feedback. This configuration demonstrates the framework's capability for high-precision manipulation tasks. The active head tracking system employs a 2-DOF gimbal that provides yaw and pitch rotation following the operator's head movements in real-time. The roll DOF is intentionally omitted to prevent motion sickness caused by inconsistency between visual and vestibular senses~\cite{de1998roll}. For stereoscopic visual feedback, a PICO 4 Ultra headset is mounted on the active head to serve as the stereo camera system that provides 2160×810 resolution stereo video streaming at 60 Hz. 

The system is validated on high-precision insertion tasks, specifically inserting a screwdriver with a 3mm diameter into a circular hole with a 4mm diameter, requiring precise spatial perception and fine motor control with only 0.5mm tolerance on each side.

\subsection{Motion Tracker for Redundant Manipulator Control}
\label{sec:motion_tracker_task}
For redundant manipulators, auxiliary motion trackers can be integrated to provide additional control. Fig.~\ref{fig:applications}(c) demonstrates the motion tracker teleoperation with a Unitree G1 upper body visualized in MeshCat\footnote[6]{\url{https://github.com/meshcat-dev/meshcat}}. The motion trackers are attached to the operator's elbows to provide position-only tracking that serves as additional inverse kinematics constraints (Eq.~\eqref{eq:constraints}) for 7-DOF arms. This elbow tracking enables intuitive control of redundant arms by resolving kinematic redundancies in an anthropomorphic manner, allowing operators to achieve more natural arm configurations while maintaining end-effector tracking from the controllers. 

\subsection{Dexterous Hand Control in MuJoCo}
Fig.~\ref{fig:applications}(d) showcases the hand pose tracking task within a MuJoCo simulation. In contrast to demonstrations that rely on XR controllers, this configuration utilizes the hand tracking mode of the headset to directly capture finger and hand gestures. The implementation employs the dexterous hand retargeting approach described in Sec.~\ref{sec:dex_retarget}, mapping the 26-joint OpenXR hand model to the Shadow Hand's kinematic structure. The hand pose tracking task demonstrates XRoboToolkit's capability to support dexterous manipulation, enabling operators to perform fine manipulation tasks through direct hand gesture control without requiring additional hardware beyond the XR headset.

% \subsubsection{Cameras}
% \begin{itemize}
%     \item Realsense RGBD cameras (435i and 405i)
% \end{itemize}

% \subsection{Simulation}
% \begin{itemize}
%     \item MuJoCo
%     \item RoboVerse
% \end{itemize}

% \subsection{Teleoperation Modes}
% \begin{itemize}
%     \item Relative end-effector position tracking
%     \item head tracking
%     \item wrist position tracker + hand tracking
% \end{itemize}
\section{Experiments}\label{sec:experiments}
\subsection{Video Streaming Latency Comparison}
\begin{figure}
    \centering
    \includegraphics[width=0.95\linewidth]{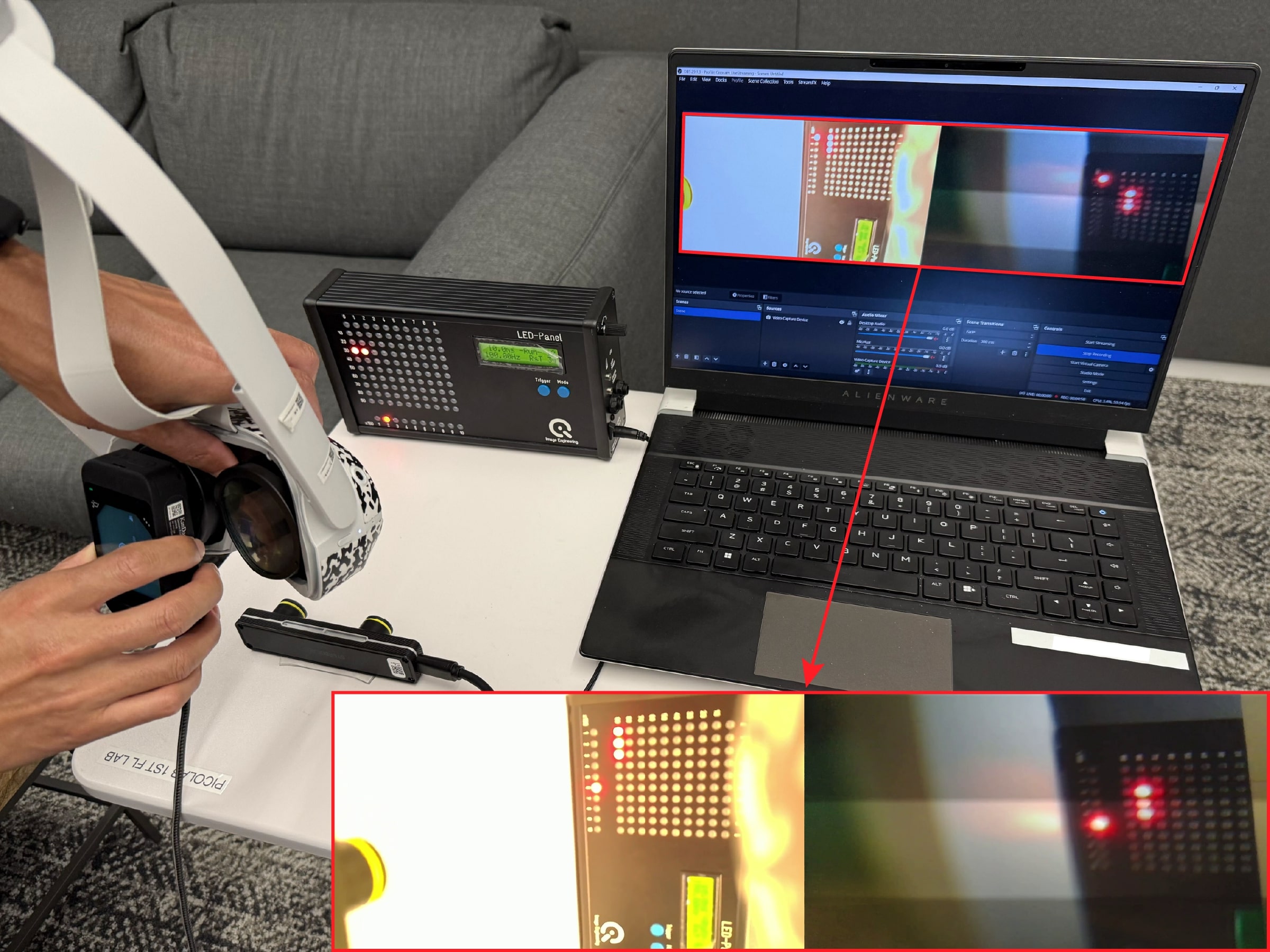}
    \caption{Video streaming latency measurement method.}
    \label{fig:measurement}
    \vspace{-0.1in}
\end{figure}

A low-latency video streaming solution was developed and comparatively evaluated against Open-TeleVision~\cite{chengopen}. To capture both virtual and real-world perspectives simultaneously, a Kandao QooCam EGO 3D Camera was used for dual-view recording. Latency measurements utilized a Precise Timing Measurement LED Panel cycling at 100 Hz. For each measurement, the illuminated LED sequence was identified, with the timestamp taken at the sequence midpoint. Latency was defined as the temporal offset between the VR display and the corresponding real-world LED panel view. Video footage was captured using OBS Studio, sampling 10 frames per condition to compute the mean and standard deviation. Fig.~\ref{fig:measurement} illustrates the measurement method and system setup.

Three conditions were evaluated: 1) Open-TeleVision, 2) ZED Mini to Quest 3, 3) ZED Mini to PICO 4 Ultra, and 3) PICO 4 Ultra to PICO 4 Ultra. For conditions 1, 2, and 3, the ZED Mini connected to a Windows 11 laptop (Intel i9-13900HK, 32 GB RAM, NVIDIA RTX 4080) using ZED SDK 4.0.8. Condition 4 used no PC; video streamed directly from the PICO 4 Ultra headset. All devices are connected to the same local network. The video transmission parameters are 1280×720 resolution, 60 FPS, and 1 Mbps bitrate, matching Open-TeleVision's default configuration.
% : 1280×720 resolution, 60 FPS, and 1 Mbps bitrate.
    
% Results are summarized in Table~\ref{tab:latency}. XRoboToolkit (ZED Mini – PICO 4 Ultra) achieved the lowest mean latency (82.00 ms) substantially outperforming Open-TeleVision (121.50 ms) and XRoboToolkit (PICO 4 Ultra – PICO 4 Ultra) at 100.50 ms. The ZED Mini – PICO 4 Ultra configuration benefits from external laptop processing, which is more computationally powerful than the standalone PICO 4 Ultra headset, likely contributing to superior latency performance. The PICO 4 Ultra – PICO 4 Ultra configuration showed the lowest variability (STD = 3.12 ms), indicating more consistent performance, while ZED Mini – PICO 4 Ultra and Open-TeleVision setups showed higher fluctuations (6.32 ms and 6.01 ms, respectively).
Results are summarized in Table~\ref{tab:latency}. Among all configurations, XRoboToolkit (ZED Mini – PICO 4 Ultra) achieved the lowest mean latency at 82.00 ms. Comparing identical hardware setups, XRoboToolkit (ZED Mini – Quest 3) at 94.50 ms substantially outperformed Open-TeleVision (ZED Mini – Quest 3) at 121.50 ms, representing a 27 ms (22\%) latency reduction. The ZED Mini – PICO 4 Ultra configuration benefits from external laptop processing, which is more computationally powerful than the standalone PICO 4 Ultra headset, likely contributing to its superior performance. The PICO 4 Ultra – PICO 4 Ultra configuration showed the lowest variability (STD = 3.12 ms), while configurations using external processing showed higher fluctuations.

\begin{table}[]
\centering
\caption{Video streaming latency comparison}
\label{tab:latency}
\begin{adjustbox}{width=\columnwidth}
\begin{tabular}{l|llll}
\toprule
\textbf{Approach} & Open-TeleVision & XRoboToolkit & XRoboToolkit & XRoboToolkit \\
\midrule
TX Device & ZED Mini & ZED Mini & ZED Mini & PICO 4 Ultra \\
RX Device & Quest 3 & Quest 3 & PICO 4 Ultra & PICO 4 Ultra \\
\midrule
Mean (ms) & 121.50 & 94.5 & 82.00 & 100.50 \\
STD (ms) & 6.01 & 7.25 & 6.32  & 3.12 \\
\bottomrule
\end{tabular}
\end{adjustbox}
\vspace{-0.2in}
\end{table}

\subsection{Data Collection for VLA Fine-tuning}
\label{sec:vla}
To validate that the teleoperation and data collection pipeline provided by XRoboToolkit can generate high-quality demonstration data suitable for VLA training, we collected 100 demonstrations of a bimanual carpet folding task using the ARX R5 dual-arm system equipped with RealSense D405i wrist cameras and a D435i overhead camera. The task sequence involves first folding the carpet in half along the short edge, then folding it again along the long edge, and finally pulling the carpet aside with the right arm. Each demonstration was recorded at 50 FPS, with every frame containing 14-dimensional robot joint states, 14-dimensional position control commands, and 424×240 RGB images from all three cameras. The average task completion time for each teleoperation demonstration was 20 seconds, with occasional regrasping and repositioning behaviors.

The dataset was used for Low-Rank Adaptation fine-tuning on the $\pi_0$ model~\cite{black2024pi_0}. Training was conducted for 80,000 steps with a batch size of 16 and an action horizon of 50 frames. The resulting policy achieved a 100\% success rate during 30 minutes of continuous operation with an average task completion time of 30 seconds. Importantly, the policy demonstrated adaptive behaviors including autonomous regrasping when grippers failed to secure the carpet and intelligent repositioning when the carpet was off-center.

\section{Conclusions}\label{sec:conclusions}
This paper presents XRoboToolkit, a cross-platform framework for XR-based robot teleoperation that addresses key limitations in existing systems through low-latency stereoscopic feedback, optimization-based control, and modular architecture. The framework demonstrates versatility across diverse robotic platforms and validates its effectiveness through precision manipulation tasks and VLA model training. While XRoboToolkit provides significant advances in accessibility and scalability, certain limitations remain. Current whole-body tracking relies on PICO's 24-joint model due to the absence of standardized whole-body definitions in OpenXR, potentially creating compatibility issues with other XR brands using different skeletal models. Furthermore, while whole-body tracking data is provided, it has not been validated through retargeting to humanoid robots for whole-body teleoperation. Additionally, the hand retargeting framework assumes each joint is individually controllable and therefore cannot accurately retarget to robot hands with mechanical constraints that couple joint movements, such as the INSPIRE Hands. The framework currently supports only MuJoCo simulation, limiting its applicability across diverse simulation environments. 

Future work will focus on improving hand retargeting algorithms for underactuated systems, expanding simulation support through platforms such as Roboverse~\cite{geng2025roboverse} to enable multi-simulator compatibility, and developing humanoid teleoperation capabilities with validated whole-body motion retargeting~\cite{ze2025twist}. Additionally, we will contribute to OpenXR standardization efforts to enable more consistent cross-platform compatibility.

\bibliographystyle{IEEEtran}
\bibliography{references.bib}

\end{document}